# COM Adjustment Mechanism Control for Multi-Configuration Motion Stability of Unmanned Deformable Vehicle

**Jun Liu\*, Hongxun Liu, Cheng Zhang, Jiandang Xing, Shang Jiang, Ping Jiang**

School of Automotive and Traffic Engineering, Hefei University of Technology, Hefei, 230009, China
Correspondence to: Jun Liu (ljun@hfut.edu.cn)

**Abstract:** An unmanned deformable vehicle is a wheel-legged robot that can transform between two configurations: a vehicular state and a humanoid state, which have different motion modes and stability characteristics. Aiming at the motion stability of an unmanned deformable vehicle in multiple configurations, a center-of-mass adjustment mechanism was designed in this study. Further, a motion stability hierarchical control algorithm was proposed based on this mechanism, and an electromechanical model based on a two-degree-of-freedom center-of-mass adjustment mechanism was established. An unmanned-deformable-vehicle vehicular-state steady-state steering dynamics model and a gait planning kinematic model of humanoid state walking were established. A stability hierarchical control strategy was designed based on the hybrid automata model, Fuzzy-PID control, K-means clustering algorithm, and variable universe fuzzy control–active disturbance rejection control (VUFC-ADRC) to realize the stability control of the unmanned deformable vehicle in multi-configuration motion. The simulation and test results showed that the steady-state steering stability in the vehicular state and the walking stability in the humanoid state could be significantly improved by controlling the slider motion in the center-of-mass adjustment mechanism.

## 1. Introduction

The unmanned deformable vehicle is a new type of deformable robot designed at Hefei University of Technology. It was designed with the precise combination of bipedal robots and electric four-wheel-drive vehicles through components such as leg mixer mechanism, horizontal lifting parallel mechanism, and COM adjustment mechanism, which is an innovative research area for wheel-legged robots. The unmanned deformable vehicle has vehicle structural features such as a chassis with a low COM height, suspension system, and wire-control steering system, allowing it to realize high-speed and stable driving in a vehicular state on structural roads. When facing non-structural roads (e.g., steps and gullies), the folding legs installed under the chassis unfold and stand up, and the unmanned deformable vehicle can be reconfigured into a bipedal robot for stride walking, thus improving the roadworthiness. The unmanned deformable vehicle has multiple motion modes to adapt to a variety of road conditions and can be applied to military fields, interstellar exploration, logistics, transportation, and other fields. At present, Hefei University of Technology has carried out in-depth research on the reconfiguration motion state analysis, reconfiguration stability control, and transient steering stability control of the vehicular state in unmanned deformable vehicles [1][2][3][4][5][6]. Based on the research described above, the stability of the unmanned deformable vehicle during walking and steady-state steering were analyzed and controlled by the COM adjustment mechanism in this study.

In traditional stability studies of the walking motion of legged robots, the zero-moment point (ZMP) is one of the most commonly used methods to determine the stability of robots [7][8]. Victor et al. [9] used the linear-quadratic regulator control algorithm to change the ZMP by adjusting the tilt angle of the robot's body to provide the wheeled bipedal Ascento robot with stronger terrain adaptability and stability in corner traveling. Zhou et al. [10] used online planning of the robot trajectory based on the ZMP stability criterion and real-time adjustment of the length of each support leg to control the position and attitude of the robot torso to solve the problem of adaptive and stable walking of quadrupedal robots on unstructured roads. Yi et al. [11] proposed a walking vibration suppression method based on optimal control for the vibration problem during the walking of a humanoid robot, introducing the vibration model into the robot dynamics model and obtaining the desired ZMP trajectory based on the optimal control theory to suppress the vibrations during the walking process and improve the walking stability. In order to optimize the dynamic stability of bipedal robot walking in real time, Kim et al. [12] generated the desired ZMP based on the capture point trajectory method and used a sliding mode controller to follow the desired ZMP so that the robot could perform various dynamic walking commands such as forward stride, lateral stride, walking direction, single support time, and double support time. Hyeok et al. established an energy-efficient gait planning and control algorithm based on a three-dimensional (3D) inverted pendulum in a stable ZMP region [13]. Nicola et al. used a linear inverted pendulum as a predictive model and ZMP as a control input to control the humanoid robot gait in real time [14].

All of the above studies were based on the traditional ZMP method for the stability control of walking legged robots. This method is a qualitative stability criterion method which uses whether the ZMP stays within the support region as the criterion for determining the system instability. The system will be destabilized once the ZMP exceeds the boundary of the support region. If this qualitative method is used for walking stability control, the stability of the ZMP cannot be quantified and graded, so the accuracy and real-time effect of the control need to be further improved. In addition, the adjustment of the ZMP in the above study was realized through the control of the robot joint motors, which needed to complete both the planned gait walking and the stability control based on ZMP adjustment when the system would be destabilized under the external disturbances. It was therefore not possible to ensure the continuity and smoothness of the walking motion in the critical destabilization state of the system. In this study, when the unmanned deformable vehicle walked in a humanoid state, the leg joints were used as the actuators for planning gait, and the COM adjustment mechanism was introduced as the actuator for walking stability to ensure the continuity and smoothness of the walking motion in the critical instability state. In terms of the walking stability criterion, the ZMP method and K-means clustering method were innovatively combined to realize the quantitative grading of the stability of the walking motion of the humanoid state, which improved the accuracy and real-time performance of the control algorithm. The variable universe fuzzy control–active disturbance rejection control (VUFC-ADRC) walking stability control algorithm was designed, which effectively reduced or shielded the external interference of the control process through the active disturbance rejection control (ADRC), as well as transformed the complex ZMP control into slider displacement control based on the COM adjustment mechanism through a variable universe fuzzy control strategy, thus reducing the complexity of the control model and facilitating the establishment of the extended state observer in the ADRC controller.

In a traditional study of vehicle steering driving stability control, Iida et al. [15] controlled the steering stability of a vehicle by direct yaw moments generated by differential braking of the four wheels. Go et al. [16] proposed an active front wheel steering control algorithm that combined a yaw rate controller with a tire force perturbation observer to coordinate the utilized attachment coefficients of the inner and outer wheels during steering, thus improving the lateral stability of the vehicle. Jalali et al. [17] proposed a stability integrated controller utilizing differential braking and active front wheel steering based on a model predictive control (MPC) algorithm in order to improve the handling stability and steering safety of the vehicle. All of the above studies used the vehicle's braking system and steering system as actuators and changed the vehicle system dynamics through the steering angle and braking torque to realize the transient steering stability control of the vehicle. However, it should be noted that very little research has been done on the steady-state steering stability control of the vehicle. The Hefei University of Technology has done in-depth research on the transient steering stability control of unmanned deformable vehicles. In order to further improve the steering performance of the vehicle, a control study on the steady-state steering stability of the vehicle was carried out by adjusting the position of the longitudinal COM of the unmanned deformable vehicle to change the dynamics of the system based on the COM adjustment mechanism.

In this study, aiming at the multi-configuration motion stability of unmanned deformable vehicle, a COM adjusting mechanism was designed and its electromechanical model was established. A multi-configuration motion stability hierarchical controller was designed based on the analysis of steady-state steering in vehicle vehicular state and gait planning in humanoid state. In the multi-configuration motion stability hierarchical controller, the upper controller was a decision controller, which classified multiple motion conditions and made stability control decisions through a hybrid automata model. The lower controller was the motion stability controller. First, a fuzzy proportional–integral–derivative (Fuzzy-PID) controller was designed to control the stability of vehicular-state steady-state steering. Second, in order to assess the grade of the walking motion stability in the humanoid state, the ZMP stability criterion was integrated with the K-means clustering method. Finally, in order to enhance the anti-interference performance of the system and build an observer, a VUFC-ADRC controller was designed to control the walking stability in the humanoid state.

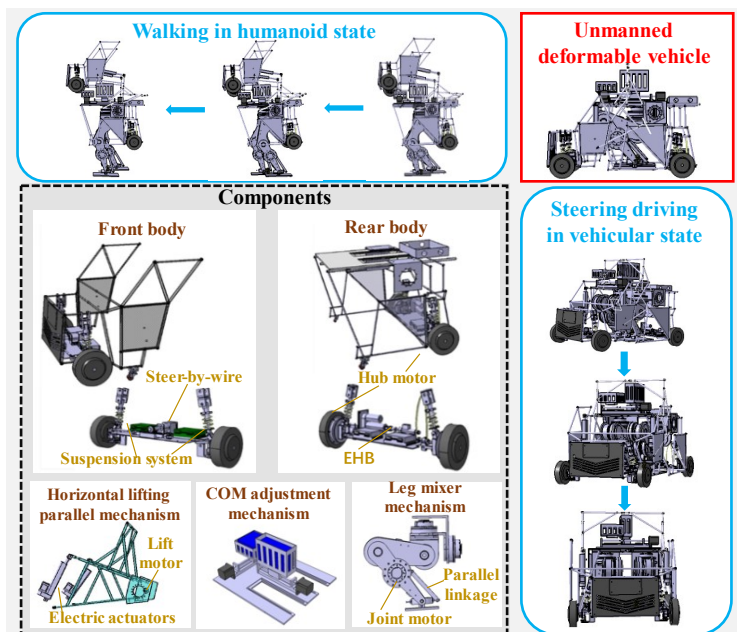

**Fig. 1.** Components and motion modes of unmanned deformable vehicle.

## 2. Structure and motion modeling of COM adjustment mechanism

*2.1. Structure and multi-configuration motion of unmanned deformable vehicle*

As shown in Fig. 1, the whole vehicle structure of the unmanned deformable vehicle is composed of a front body, a rear body, a COM adjustment mechanism, a horizontal-lifting parallel mechanism, and a leg mixer mechanism. The robot can be deformed between two configurations, the vehicular state and the humanoid state, and has two corresponding movement modes of driving and walking, respectively. The front body is connected to the rear body through the horizontal-lifting parallel mechanism. When the unmanned deformable vehicle is reconfigured, the horizontal-lifting of the front body is realized by controlling the motor and electric actuator in the horizontal-lifting parallel mechanism. The leg mixer mechanism was mounted on the rear body, and each leg had six degrees of freedom. In order to reduce the rotational inertia of the legs, the ankle joint motors are moved up to the middle of the calf and driven by a parallel linkage mechanism.

When the unmanned deformable vehicle is in the vehicular state, the leg mixer mechanism is folded and stowed in the vehicle chassis. The unmanned deformable vehicle is driven, steered, and braked by four hub motors, steer-by-wire, and a braking system, respectively, to allow for high-speed driving on a structured road. When the unmanned deformable vehicle is in the humanoid state, the leg mixer mechanism is deployed and stands on the ground. The front body is lifted to become the upper body of the humanoid state, and the unmanned deformable vehicle is driven by the leg joint motors to complete walking according to the planned gait.

*2.2. Structural design of COM adjustment mechanism*

The COM adjustment mechanism is an innovative component applied to the unmanned deformable vehicle, in which the COM position is adjusted by the movement of the slider in the mechanism to control the stability of the motion in the vehicular and humanoid states. The COM adjustment mechanism is bolted to the roof plate of the body, as shown in Fig. 2, and consists of X- and Y-direction slides, a battery, and a drive motor. The battery functions both as an adjustable COM slider and as a power source for the unmanned deformable vehicle. The X- and Y-direction slides are spatially arranged in a T-shape to avoid cross-interference of movements. The output shaft of the motor is connected to a ball screw through a coupling, and the rotation of the motor is converted into the movement of the slider on the slide rail by the ball screw. When the unmanned deformable vehicle performs steady-state steering in the vehicular state, the most ideal understeering state is maintained by controlling the movement of the X-direction slider. When the unmanned deformable vehicle is walking in the humanoid state, the single-legged and double-legged supports will appear in turn, and the support area changes noticeably. Therefore, it is necessary to jointly control the X- and Y-direction sliders in the COM adjustment mechanism to ensure the stability of walking in the humanoid state.

*2.3. Electromechanical model of COM adjustment mechanism*

The current and torque loops of the COM adjustment mechanism are equated to a current constant $K_a$ and a torque constant $K_t$, respectively, and the output torque $T_t$ of the motor is

$$T_t = K_a K_t u, \qquad (1)$$

where $u$ is the output voltage. The dynamics of the motor are modeled as

$$J\ddot{\theta} + B_z \dot{\theta} = T_t, \qquad (2)$$

where $J$ and $B_z$ are the equivalent moment of inertia and the equivalent viscous damping coefficient of the COM adjustment mechanism, respectively, and $\theta$ is the motor rotation angle.

The displacement $y$ of the adjusted COM slider on the slide is

$$y = r_g \theta, \qquad (3)$$

where $r_g$ is the transmission ratio of the linear guide rod, $r_g = p_r/2\pi$, and $p_r$ is the lead of the guide rod.

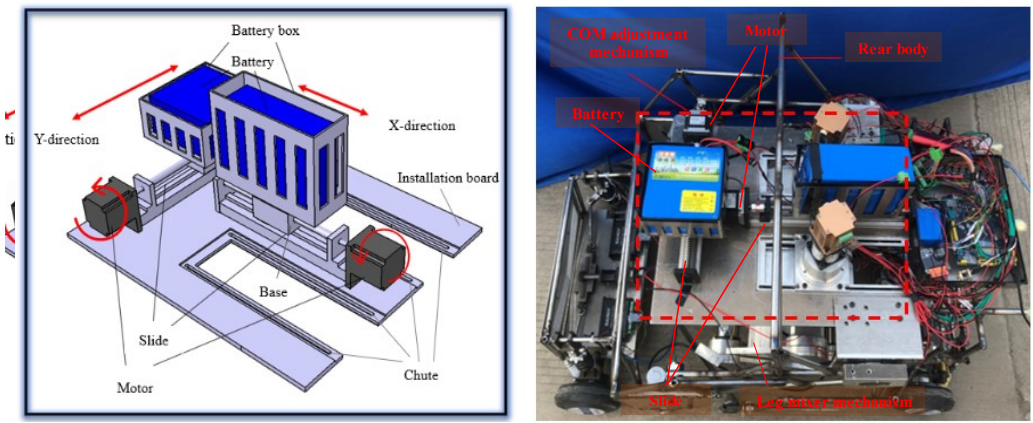

**Fig. 2.** COM adjustment mechanism.

By combining Eqs. (1)–(3), the following expression is obtained for the mechatronic model of the COM adjustment mechanism:

$$\ddot{y} = -\frac{B_z}{J}\dot{y} + \frac{K_a K_t r_g}{J}u. \tag{4}$$

## 3. Multi-configuration motion analysis

### 3.1. Vehicular-state steady-state steering

As shown in Fig. 3, when the unmanned deformable vehicle is traveling via steady-state steering, its lateral and yaw two-degree-of-freedom differential equations are

$$\begin{cases} (k_1 + k_2)\beta + \dfrac{1}{u_x}(ak_1 - bk_2)\omega_r - k_1\delta = m(\dot{u}_y + u_x\omega_r) \\ (ak_1 - bk_2)\beta + \dfrac{1}{u_x}(a^2 k_1 - b^2 k_2)\omega_r - ak_1\delta = I_z\dot{\omega}_r \end{cases}, \tag{5}$$

where $m$ is the total mass of the unmanned deformable vehicle, $k_1$ and $k_2$ are the camber aligning stiffness coefficients of the front and rear wheels, respectively, $a$ represents the distance from the COM to the front axle, $b$ is the distance from the COM to the rear axle, $\omega_r$ is the yaw rate, $I_z$ is the rotational inertia of the unmanned deformable vehicle around the z-axis, $u_x$ is the longitudinal velocity, $u_y$ is the lateral velocity, $\beta$ is the slip angle, and when the slip angle is smaller, $\beta = u_y/u_x$, and $\delta$ is the front wheel steering angle.

When the unmanned deformable vehicle drives at a constant speed, $\dot{u}_y = 0$ and $\dot{\omega}_r = 0$, which are introduced into the above equation, yielding the following:

$$\begin{cases} (k_1 + k_2)\beta + \dfrac{1}{u_x}(ak_1 - bk_2)\omega_r - k_1\delta = mu_x\omega_r \\ (ak_1 - bk_2)\beta + \dfrac{1}{u_x}(a^2 k_1 - b^2 k_2)\omega_r - ak_1\delta = 0 \end{cases}. \tag{6}$$

When the unmanned deformable vehicle drives at a constant speed, $\dot{u}_y = 0$ and $\dot{\omega}_r = 0$, which are introduced into the above equation, yielding the following:

$$\begin{cases} (k_1 + k_2)\beta + \dfrac{1}{u_x}(ak_1 - bk_2)\omega_r - k_1\delta = mu_x\omega_r \\ (ak_1 - bk_2)\beta + \dfrac{1}{u_x}(a^2 k_1 - b^2 k_2)\omega_r - ak_1\delta = 0 \end{cases}. \tag{6}$$

The yaw rate gain can be obtained by combining Eq. (5) and Eq. (6):

$$\left.\frac{\omega_r}{\delta}\right)_s = \frac{u_x/L}{1 + \dfrac{m}{L}\left(\dfrac{a}{k_2} - \dfrac{b}{k_1}\right)u_x^2} = \frac{u_x/L}{1 + Ku_x^2}, \tag{7}$$

where $L$ is the wheelbase in the vehicular state, and $K$ is the stability factor, which is an important parameter used to characterize the steady-state steering. Its formula is as follows:

$$K = \frac{m}{L^2}\left(\frac{a}{k_2} - \frac{b}{k_1}\right). \tag{8}$$

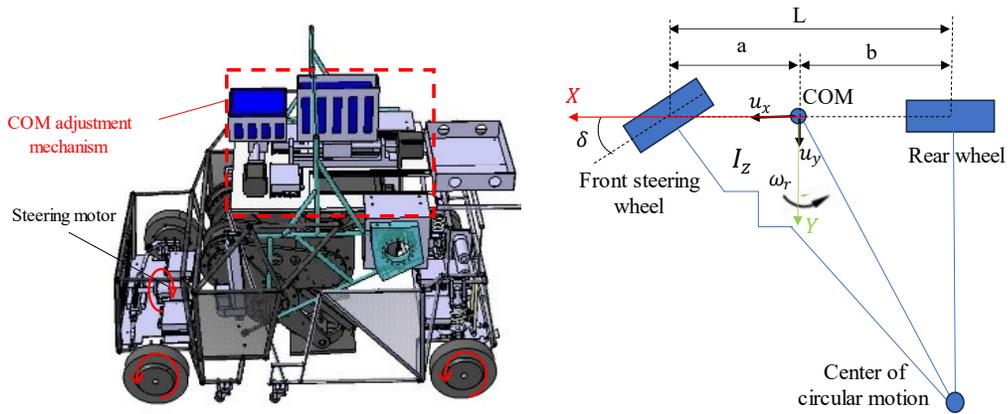

**Fig. 3.** Steady-state steering model of unmanned deformable vehicles.

The yaw rate gain $\left.\dfrac{\omega_r}{\delta}\right)_s$ and the stability factor $K$ are the key parameters for evaluating the steady-state steering performance of a vehicle [18]. When $K = 0$, $\left.\dfrac{\omega_r}{\delta}\right)_s = u_x/L$, the yaw rate gain is linearly related to the vehicle speed, and the steady-state steering behaves as neutral steering. When $K < 0$, the steady-state steering behaves as over-steering, and the yaw rate gain increases exponentially with the increase in the vehicle speed. When the vehicle speed $u_x = \sqrt{-1/K}$, the yaw rate tends to infinity, which means that the vehicles destabilize with minor steering angle. When $K > 0$, the steady-state steering of the vehicular state manifests itself as understeering, and the yaw rate gain will rapidly increase before plateauing as the vehicle speed increases. Therefore, the vehicular state should have a moderate understeering ability to avoid the vehicle steering instability due to the excessive yaw rate gain. The X-direction COM position $C_x$ of the unmanned deformable vehicle can be adjusted in real time by the COM adjustment mechanism, which changes the front wheelbase $a$, the rear wheelbase $b$, and the stability factor $K$ of the vehicular state to ensure that the vehicle maintains a good understeering ability at all times.

*3.2. Humanoid walking*

The kinematic model of the unmanned deformable vehicle during walking is established using the homogeneous coordinate method. As shown in Fig. 4, the base coordinate system of the unmanned deformable vehicle in humanoid state is $O_0 - X_0Y_0Z_0$, the coordinate origin is set at the centers of the two legs, the positive direction of the X-axis is the forward direction, the positive direction of the Z-axis is vertically upward on the ground, and the positive direction of the Y-axis is determined by the right hand rule. The attachment coordinate system of the ankle, knee, and hip joints of the right leg and the hip, knee, and ankle joints of the left leg are denoted sequentially as $O_i - X_iY_iZ_i (i = 1 \ldots 6)$, and the coordinate system of the lifting joints is $O_7 - X_7Y_7Z_7$. The unmanned deformable vehicle can be divided into a single-legged support period and a double-legged support period during the walking process.

The angle and translation distance of the coordinate system $O_i - X_iY_iZ_i$ relative to $O_{i-1} - X_{i-1}Y_{i-1}Z_{i-1}$ along the X-axis, Y-axis, and Z-axis are defined to be $\theta_{ix}$, $\theta_{iy}$, $\theta_{iz}$, $l_{x(i-1)}$, $l_{y(i-1)}$, and $l_{z(i-1)}$, respectively. The homogeneous coordinate transformation matrix $^{i-1}_iT$ $(i = 1 \ldots 7)$ between neighboring coordinate systems can be expressed as

$$^{i-1}_iT = \begin{pmatrix} cos\theta_{iz}cos\theta_{ix} & -cos\theta_{iz}cos\theta_{ix} & sin\theta_{iz} & l_{x(i-1)} \\ sin\theta_{iy}c_{iy} + cos\theta_{ix}sin\theta_{iy}cos\theta_{iz} & cos\theta_{iy}^2 - sin\theta_{iy}sin\theta_{iz}sin\theta_{ix} & -cos\theta_{iz}sin\theta_{iy} & l_{y(i-1)} \\ cos\theta_{iy}^2 - cos\theta_{iy}cos\theta_{ix}sin\theta_{iz} & sin\theta_{iy}cos\theta_{iy} + cos\theta_{iy}sin\theta_{iz}sin\theta_{ix} & cos\theta_{iy}cos\theta_{iz} & l_{z(i-1)} \\ 0 & 0 & 0 & 1 \end{pmatrix}. \tag{9}$$

The homogeneous coordinate transformation matrix of any attached coordinate system with respect to the base coordinate system can be expressed as

$$^0_iT = ^0_1T \ldots ^{i-1}_iT. \tag{10}$$

By knowing the COM vector $c_i = [c_{xi}, c_{yi}, c_{zi}]$ of each component in its own attachment coordinate system, the COM vector $c_i'$ of each component in the base coordinate system can be obtained:

$$c_i' = [c_{xi}', c_{yi}', c_{zi}', 1]^T = ^0_iT[c_{xi}, c_{yi}, c_{zi}, 1]^T. \tag{11}$$

The velocity $\dot{c}_i'$ and acceleration $\ddot{c}_i'$ of each component can be obtained by taking the first- and second-order derivatives of the above equation, respectively:

$$\dot{c}_i' = [\dot{c}_{xi}', \dot{c}_{yi}', \dot{c}_{zi}'], \tag{12}$$
$$\ddot{c}_i' = [\ddot{c}_{xi}', \ddot{c}_{yi}', \ddot{c}_{zi}']. \tag{13}$$

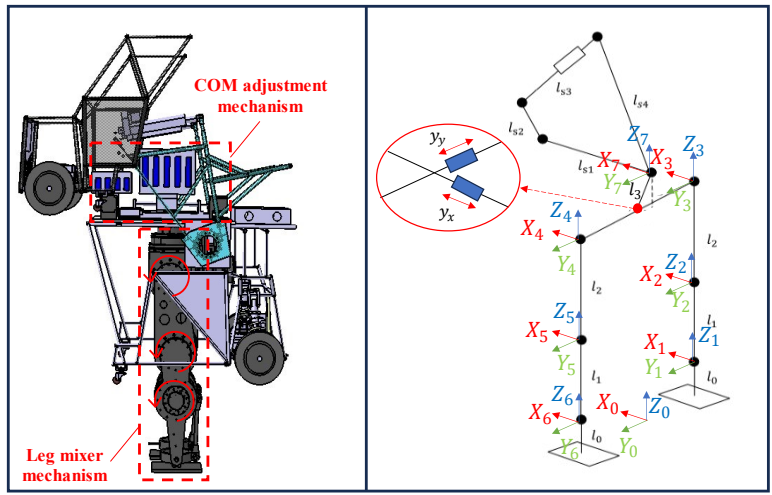

**Fig. 4.** Kinematic modeling of unmanned deformable vehicle walking in humanoid state.

The mass of each component is $m_i$ and the formulas for the coordinates of the COM position of the unmanned deformable vehicle during walking can be obtained as

$$\begin{cases} C_x = \frac{\sum_{i=1}^n m_i c_{xi}'}{m} \\ C_y = \frac{\sum_{i=1}^n m_i c_{yi}'}{m} \\ C_z = \frac{\sum_{i=1}^n m_i c_{zi}'}{m} \end{cases} \tag{14}$$

The ZMP method is used as a criterion for judging the stability of the walking motion, and the expression for ZMP can be obtained as follows:

$$\begin{cases} ZMP_x = \frac{\sum_{i=1}^n m_i(\ddot{c}_{zi}' + g)c_{xi}' - \sum_{i=1}^n m_i \ddot{c}_{xi}' c_{zi}'}{\sum_{i=1}^n m_i(\ddot{c}_{zi}' + g)} \\ ZMP_y = \frac{\sum_{i=1}^n m_i(\ddot{c}_{zi}' + g)c_{yi}' - \sum_{i=1}^n m_i \ddot{c}_{yi}' c_{zi}'}{\sum_{i=1}^n m_i(\ddot{c}_{zi}' + g)} \end{cases}, \tag{15}$$

where $g$ is the gravitational acceleration.

While walking, an inverted pendulum model is used to plan the gait [19]. The unmanned deformable vehicle walking comprises multiple stride cycles (each stride cycle contains two single-legged support periods $T_s$ on the left and right and the corresponding two double-legged support periods $T_d$). The height of the COM of the unmanned deformable vehicle during the walking process is 0.9 m, the stride length is 0.4 m, and the stride height is 0.1 m. During the single-legged support period, only the ankle, knee, and hip motors for the leg around the Y-direction and the COM adjusting sliders for the X- and Y-directions are the active parts. The coordinate trajectory of the hip joint is planned using the inverted pendulum model, and the coordinate trajectory of the ankle joint is planned using the displacement fifth-degree polynomial. The angle of each joint is solved for sequentially by the inverse kinematics, as follows:

$$\begin{cases} \theta_1 = arccos\left(\frac{l_{l1l3}^2 + l_1^2 - l_2^2}{2l_1 l_{l1l3}}\right) + arctan\left(\frac{x_{l3} - x_{l1}}{z_{l3} - z_{l1}}\right) \\ \theta_2 = -arccos\left(\frac{l_2^2 + l_1^2 - l_{l1l3}^2}{2l_1 l_2}\right) \\ \theta_3 = \pi - \theta_1 - arccos\left(\frac{l_2^2 + l_1^2 - l_{l1l3}^2}{2l_1 l_2}\right) \\ \theta_4 = \pi - \theta_6 - arccos\left(\frac{l_{r1r3}^2 + l_1^2 - l_2^2}{2l_1 l_{r1r3}}\right) \\ \theta_5 = -arccos\left(\frac{l_1^2 + l_2^2 - l_{r1r3}^2}{2l_1 l_{r1r3}}\right) \\ \theta_6 = arccos\left(\frac{l_{r1r3}^2 + l_1^2 - l_2^2}{2l_1 l_{r1r3}}\right) - arctan\left(\frac{x_{r1} - x_{r3}}{z_{r3} - z_{r1}}\right) \end{cases}, \tag{16}$$

where $\theta_i (i = 1, ..., 6)$ are the ankle, knee, and hip joints of the right leg and the hip, knee, and ankle joints of the left leg, respectively, and $l_1$ and $l_2$ are the calf and thigh rod lengths, respectively. The coordinates of the hip joint of the left leg are ($x_{l3}$, $z_{l3}$), the coordinates of the ankle joint are ($x_{l1}$, $z_{l1}$), the coordinates of the hip joint of the right leg are ($x_{r3}$, $z_{r3}$), and the coordinates of the ankle joint are ($x_{r1}$, $z_{r1}$). The distance between the hip and ankle joints of the left leg $l_{l1l3} = \sqrt{(x_{l3} - x_{l1})^2 + (z_{l3} - z_{l1})^2}$, and that between the hip and ankle joints of the right leg $l_{r1r3} = \sqrt{(x_{r3} - x_{r1})^2 + (z_{r3} - z_{r1})^2}$. By simulating Eq. (16), the motion trajectory of each joint in each stride cycle is obtained, as shown in Fig. 5.

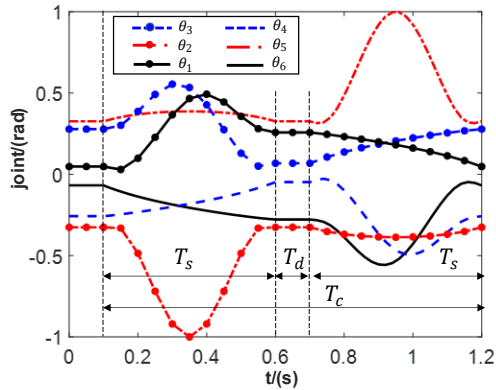

**Fig. 5.** Motion trajectory of leg joint motor.

# 4. Multi-configuration motion stability control strategy based on COM adjustment mechanism

The unmanned deformable vehicle may be destabilized when walking in the humanoid state or during steady-state steering in the vehicular state. In order to ensure the motion stability of the unmanned deformable vehicle during walking or steady-state steering, the motion stability control strategy designed based on the COM adjustment mechanism is illustrated in Fig. 6. The multi-configuration motion stability control strategy based on the COM adjustment mechanism consists of an upper-level decision controller and a lower-level motion stability controller. The upper-level controller is based on the hybrid automata model to judge the motion conditions of the unmanned deformable vehicle according to its motion parameters and select the corresponding motion stability control strategy. The lower-level controller comprises a steady-state steering stability control module in the vehicular state and a walking stability control module in the humanoid state.

Steady-state steering stability control is realized by the Fuzzy-PID controller. The deviation $K_e$ and the rate of change of the deviation $K_{ec}$ between the desired stability factor $K_d$ and the actual stability factor $K$ are taken as the inputs of the fuzzy controller, and the three control parameters $\Delta k_p$, $\Delta k_i$, and $\Delta k_d$ of the proportional–integral–derivative (PID) controller are output after the fuzzy operation. The PID controller outputs a control voltage $u_x$ into the electromechanical model to adjust the displacement $y_x$ of the COM slider according to the deviation $K_e$. Based on the change of $y_x$, the actual stability factor $K$ of the unmanned deformable vehicle is calculated. Finally, the deviation between the actual stability factor $K$ and the desired stability factor $K_d$ is input into the Fuzzy-PID controller to complete the closed-loop feedback control.

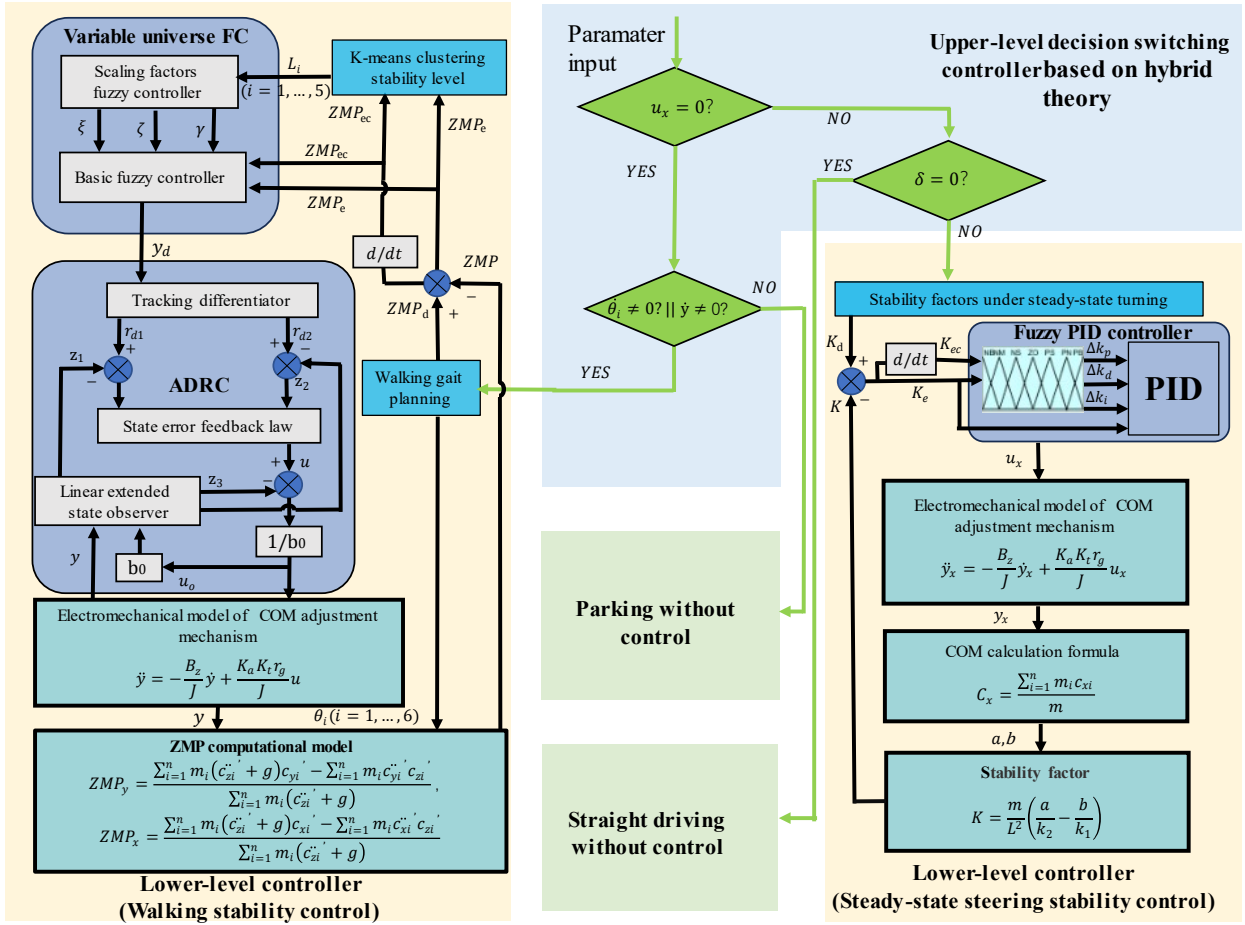

**Fig. 6.** Control block diagram of multi state motion stability of unmanned deformable vehicle.

While walking, the support domain of the humanoid state is small and varies greatly, and the change of the COM position and external disturbances may cause walking tipping instability. In the humanoid walking stability control, to suppress the influence of interference on the control system, the computationally complex desired ZMP tracking control is first transformed into the desired displacement control of the X- and Y-direction sliders of the COM adjustment mechanism through the variable universe fuzzy controller, and the tracking control of the desired displacement of the slider is subsequently completed by the ADRC. The humanoid walking stability control process is divided into the following steps:

(1) Based on the K-means clustering algorithm, the walking stability grade is quantitatively evaluated according to the deviation $ZMP_e$ and deviation change rate $ZMP_{ec}$ between the expected $ZMP_d$ and the actual $ZMP$.

(2) The evaluation result $L_i$ is input into the variable universe fuzzy controller, and the scaling factor fuzzy controller outputs

three scaling factors according to the walking stability grade, which in turn optimizes the universe of the basis fuzzy controller in real time. The basis fuzzy controller then outputs the desired slider displacement $y_d$ to the ADRC after fuzzy operation based on the optimized universe, deviation $ZMP_e$, and deviation change rate $ZMP_{ec}$.

(3) In order to track the desired slider displacement $y_d$, the ADRC first smooths the desired displacement by the tracking differentiator, then predicts the output and disturbance quantities by the expansion state observer, and finally outputs the voltage $u_0$ to the electromechanical model to obtain the actual displacement $y$ of the slider under the operation of the state error feedback.

(4) According to the actual displacement of the slider, the actual $ZMP$ of the unmanned deformable vehicle is calculated by the $ZMP$ model. Then, the deviation difference between the actual $ZMP$ and the desired $ZMP_d$ is calculated and input to the variable universe fuzzy controller to complete the closed-loop feedback control.

*4.1. Upper-level decision controller design*

A hybrid automata model is a dynamic system formed by continuous time variables and discrete events mixing and interacting with each other [20][21]. For the multi-configuration motion of the unmanned deformable vehicle, the operating parameters in each configuration (vehicle speed $u_x$, front wheel angle $\delta$, slider displacement $y$, and joint angle $\theta_i$) all show continuous change characteristics with time.

In contrast, the motions in different configurations form a series of discrete motion events. Therefore, the motion of the unmanned deformable vehicle in multiple configurations is a hybrid dynamic system. When the unmanned deformable vehicle monitors the change of the motion state, the upper-level decision controller is driven to select the corresponding control strategy according to the discrete event characteristics to carry out the motion stability control in the corresponding configuration.

The lower-level motion stability control includes steady-state steering stability control $Q_1$ in the vehicular state and walking stability control $Q_2$ in the humanoid state. In addition to the two control states, there also exist the straight-line driving state $Q_3$ and the parked state $Q_4$, which are not controlled. The steering angle $\delta$, vehicle speed $u_x$, slider displacement velocity $\dot{y}$, and joint motor angular velocity $\dot{\theta}_i$ are used to determine the motion state of the unmanned deformable vehicle. $ST\delta = 1$ represents $\delta \neq 0$, $ST\delta = 0$ represents $\delta = 0$, $STu_x = 1$ represents $u_x \neq 0$, $STu_x = 0$ represents $u_x = 0$, $ST\dot{\theta}_i = 1$ represents $\dot{\theta}_i \neq 0$, $ST\dot{\theta}_i = 0$ represents $\dot{\theta}_i = 0$, $ST\dot{y} = 1$ represents $\dot{y} \neq 0$, and $ST\dot{y} = 0$ represents $\dot{y} = 0$.

The upper-level controller based on the hybrid automata model is represented by the following eight tuples:
$$H = (X, Q, U, Y, Init, Inv, S, G).$$

(1) $X$ is a state space of continuous dynamic parts, $X = (\delta, u_x, \dot{\theta}_i, \dot{y})$.
(2) $Q$ is a set of discrete operating modes, $Q = \{Q_1, Q_2, Q_3, Q_4\}$.
(3) $U$ is a set of input variables, $U = (ST\delta, STu_x, ST\dot{\theta}_i, ST\dot{y})$.
(4) $Y$ is a set of output variables, $Y = Q$.
(5) $Init$ is the set of initial states, $Init = (Q_4, \delta = 0, u_x = 0, \dot{\theta}_i = 0, \dot{y} = 0)$.
(6) $Inv$ is the set of invariants specific to the whole system,

$$Inv(Q) = \begin{cases} Q_1 = \{(x,u) \in X \times U : [ST\delta = 1 \wedge STu_x = 1 \wedge ST\dot{\theta}_i = 0 \wedge ST\dot{y} = 1]\}, \\ Q_2 = \{(x,u) \in X \times U : [ST\delta = 0 \wedge STu_x = 0 \wedge ST\dot{\theta}_i = 1 \wedge ST\dot{y} = 1]\}, \\ Q_3 = \{(x,u) \in X \times U : [ST\delta = 0 \wedge STu_x = 1 \wedge ST\dot{\theta}_i = 0 \wedge ST\dot{y} = 0]\}, \\ Q_4 = \{(x,u) \in X \times U : [ST\delta = 0 \wedge STu_x = 0 \wedge ST\dot{\theta}_i = 0 \wedge ST\dot{y} = 0]\}. \end{cases}$$

(7) $S$ represents the transition between two discrete states, and the switching relation is shown in Fig. 7.
$S = S^{13}, S^{31}, S^{24}, S^{42}, S^{34}, S^{43}, S^{41}, S^{14} = \{(Q_1, Q_3), (Q_3, Q_1), (Q_2, Q_4), (Q_4, Q_2), (Q_3, Q_4), (Q_4, Q_3), (Q_4, Q_1), (Q_1, Q_4)\}$.

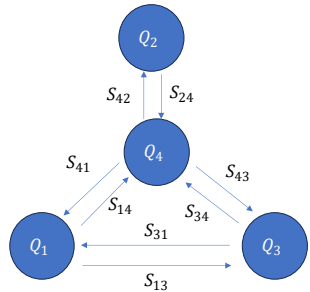

**Fig. 7.** Discrete state switching relationship.

(8) $G$ is the guarding condition. When the state change of the system is in the following the region, the switching system is activated, and the state of the system will jump:
$$G(S^{13}) = \{X \times U : ST\delta = 0\},$$
$$G(S^{31}) = \{X \times U : ST\delta = 1 \wedge ST\dot{y} = 1\},$$
$$G(S^{24}) = \{X \times U : ST\dot{\theta}_i = 0\},$$

$$G(S^{42}) = \{X \times U: ST\dot{\theta}_l = 1 \wedge ST\dot{y} = 1\},$$
$$G(S^{34}) = \{X \times U: STu_x = 0\},$$
$$G(S^{43}) = \{X \times U: STu_x = 1\},$$
$$G(S^{14}) = \{X \times U: STu_x = 0\}$$
$$G(S^{41}) = \{X \times U: STu_x = 1 \wedge ST\delta = 1 \wedge ST\dot{y} = 1\}.$$

*4.2. Lower-level motion stability controller design*

1) Stability grade discrimination based on K-means clustering

ZMP is a commonly used method for qualitatively judging stability during robot walking, but it cannot quantitatively grade the stability of the robot. In this paper, ZMP is combined with the K-means clustering algorithm to realize the quantitative grading of the stability of the unmanned deformable vehicle during walking. Different cluster centers output by the clustering algorithm [22][23][24] represent different stability levels, and the current stability level is obtained by calculating the minimum distance between the current stability parameters and each cluster center. The variable universe fuzzy controller adaptively changes the control universe and control precision according to the stability level in the walking process, so as to improve the safety.

To obtain the stability clustering dataset, the 3D model of the unmanned deformable vehicle was imported into ADAMS to simulate its walking stability changes by offline simulation. The unmanned deformable vehicle walked 10,000 steps in the simulation software and sampled each step 10 times. Finally, a total of $10^5$ actual ZMP sampling data were obtained. The ideal $ZMP$ corresponding to the sampled data was solved based on the designed walking gait, and the stability clustering dataset matrix $D$ was obtained by calculating the deviation vector $\overrightarrow{ZMP_e}$ and the deviation change rate vector $\overrightarrow{ZMP_{ec}}$ between the ideal value and the actual value of all sampling points as follows:

$$\begin{cases} D = [\overrightarrow{ZMP_e}, \overrightarrow{ZMP_{ec}}] \\ \overrightarrow{ZMP_e} = (ZMP_{e_1}, ZMP_{e_2}, \cdots ZMP_{e_{10^5}})^T \\ \overrightarrow{ZMP_{ec}} = (ZMP_{ec_1}, ZMP_{ec_2}, \cdots ZMP_{ec_{10^5}})^T \end{cases}. \tag{17}$$

The walking stability clustered dataset was divided into five levels by the K-means clustering algorithm, and a larger value of the level represented worse walking stability. The clustering process included the following steps:

(1) Five data points were randomly selected in the dataset as the initial cluster centers $(ZMP_{eoj}, ZMP_{ecoj})(j = 1,2, \cdots 5)$ of the five clusters, and the maximum number of iterations was set to 1000.

(2) For each data point $(ZMP_{e_i}, ZMP_{ec_i})(i = 1,2, \cdots 10^5)$ in the dataset, the distance to each cluster center was calculated and each data segment was assigned to the closest cluster. The weighted distance from the sample point to each cluster center was calculated as

$$\rho = \sqrt{0.7(ZMP_{e_i} - ZMP_{eoj})^2 + 0.3(ZMP_{ec_i} - ZMP_{ecoj})^2}. \tag{18}$$

(3) The average of all samples points in each cluster was calculated, and this average was used as the new cluster center.

(4) Steps (2) and (3) are repeated until the cluster center no longer changed significantly or the maximum number of iterations was reached.

By performing K-means clustering operations offline on a large amount of walking simulation data, the relationship between the stability level $L_i$ in the X-direction and the corresponding cluster center $(|ZMP_{ex}|, |ZMP_{ecx}|)$, as well as the relationship between the stability level $L_i$ in the Y-direction and the corresponding cluster center $(|ZMP_{ey}|, |ZMP_{ecy}|)$, were obtained and are shown in Table 1.

2) Fuzzy proportional–integral–derivative (PID) controller for steady-state steering stability control and variable universe fuzzy controller for walking stability control

The control parameters of the conventional PID controller are fixed, so the robustness is poor. Therefore, the steady-state steering stability control uses a fuzzy PID controller as a position controller for adjusting the COM slider. The fuzzy controller adjusts the three control parameters $\Delta k_p$, $\Delta k_i$, and $\Delta k_d$ of the PID controller in real time according to the deviation $K_e$ and the rate of change of the deviation $K_{ec}$ of the stability factor K, so that the PID controller can output the optimal control voltage $u_x$ to control the displacement of the X-direction adjusting COM slider. The front and rear wheelbases of the unmanned deformable vehicle are altered in real time to ensure that the vehicle always has the optimal stability factor K. All the fuzzy sets of the fuzzy PID controller are set as {Negative Big (NB), Negative Medium (NM), Negative Small (NS), Zero (ZO), Positive Small (PS), Positive Medium (PM), Positive Big (PB)}, and triangular affiliation functions are chosen for both input and output affiliation functions. The universes of the input variables $K_e$ and $K_{ec}$ are [−0.1,0.1] and [−0.05,0.05], respectively, and the universes of the output variables $\Delta k_p$, $\Delta k_i$, and $\Delta k_d$ are [−15,15], [−1,1], and [−15,15], respectively. The control rules of the fuzzy PID controller are shown in Table 2. The initial parameters of the fuzzy PID were selected after several debugging processes as $k_{p0} = 15$, $k_{i0} = 0.8$, and $k_{d0} = 9$.

Table 1 Clustered centers and stability classes during walking

| Level | Meaning | $|ZMP_{ex}|$ | $|ZMP_{ecx}|$ | $|ZMP_{ey}|$ | $|ZMP_{ecy}|$ |
|---|---|---|---|---|---|
| $L_1$ | Very stable | 0.0079 | 0.0195 | 0.0152 | 0.00405 |
| $L_2$ | Comparatively stable | 0.0315 | 0.0293 | 0.0213 | 0.01313 |
| $L_3$ | Sub-stable | 0.0611 | 0.0411 | 0.0352 | 0.0252 |
| $L_4$ | Critical stable | 0.0937 | 0.0506 | 0.0457 | 0.0406 |
| $L_5$ | Destabilized | 0.1149 | 0.0586 | 0.0607 | 0.0506 |

Table 2 Fuzzy rules for $\Delta k_p$, $\Delta k_i$, $\Delta k_d$

| | $\Delta k_p/\Delta k_i/\Delta k_d$ | $K_e$ | | | | | | |
|---|---|---|---|---|---|---|---|---|
| | | NB | NM | NS | ZE | PS | PM | PB |
| $K_{ec}$ | NB | PB/NB/PS | PB/NB/NS | PM/NM/NS | PM/NM/NS | PS/NS/NM | PS/ZE/NM | ZE/ZE/PM |
| | NM | PB/NB/PS | PB/NB/NS | PM/NM/NS | PM/NM/NM | PS/NS/NM | ZE/ZE/NS | ZE/ZE/ZE |
| | NS | PM/NB/ZE | PM/NM/NS | PS/NS/NM | PS/NS/NM | ZE/ZE/NM | NS/PM/NS | NS/PS/ZE |
| | ZE | PM/NM/ZE | PM/NM/NB | PS/NS/NS | ZE/ZE/NS | NS/PS/PS | NM/PM/NS | NM/PS/ZE |
| | PS | PS/NM/ZE | PS/NS/ZE | ZE/ZE/ZE | NS/PS/ZE | NS/ZE/PS | NM/PM/ZE | NM/PB/PM |
| | PM | PS/ZE/PB | ZE/ZE/PS | NS/PS/PS | NM/PS/PS | NM/PM/PS | NM/PB/PS | NB/PS/PB |
| | PB | ZE/ZE/PB | ZE/ZE/PM | NS/PS/PM | NM/PM/PM | NM/PM/PS | NB/PB/PS | NB/PS/PB |

The ranges of the universes and the number of fuzzy rule divisions have a large impact on the control effect of a fuzzy controller. When the number of fuzzy rules is larger and the range of the universe is smaller, the classification will be finer, and the control accuracy of the controller will be higher. However, with a finer class division, the solution speed of the controller will be reduced, and the negative role of measurement error in the classification will be greater. In order to improve the adaptability and accuracy of the fuzzy controller without increasing the number of fuzzy rules, a variable universe fuzzy controller is used to control the walking stability of the unmanned deformable vehicle.

The variable universe fuzzy controller consists of a basis fuzzy controller and a scaling factor fuzzy controller. The two input variables of the basis fuzzy controller are the deviation $ZMP_e$ and the deviation change rate $ZMP_{ec}$, and the output variable is the desired slider displacement $y_d$. The fuzzy sets of the input variables $ZMP_e$ and $ZMP_{ec}$ and the output variable $y_d$ are defined as {Negative Big (NB), Negative Medium (NM), Negative Small (NS), Zero (ZE), Positive Small (PS), Positive Medium (PM), Positive Big (PB)}. The input variable of the scaling factor fuzzy controller is the stability level $L_i$ of the unmanned deformable vehicle while walking, the universe is [0,1], and the fuzzy sets are defined as {Very Small ($L_1$), Small ($L_2$), Medium ($L_3$), Big ($L_4$), Very Big ($L_5$)}. The output variables of the scaling factor fuzzy controllers are the scaling factors $\zeta(ZMP_e)$, $\xi(ZMP_{ec})$, and $\gamma(y_d)$, all with universes of [0.5,1.5], and fuzzy sets defined as {Very Small (VS), Small (S), Medium (M), Big (B), Very Big (VB)}. The triangular affiliation function was chosen for all the fuzzy controllers. In order to achieve good control results, when the walking stability level was low, the scaling factor fuzzy controller output a smaller scaling factor so that the basis fuzzy controller universe was divided more finely, which ultimately improved the control accuracy and reduced the amount of overshoot. As the walking stability level became larger, the unmanned deformable vehicle faced the risk of tipping destabilization. At this time, the scaling factor fuzzy controller output a larger scaling factor to increase the control variables output by the basis fuzzy controller to realize the rapid adjustment of the ZMP. As shown in Fig. 8, the input–output initial fuzzy universe [−1,1] of the basis fuzzy controller undergoes adaptive expansion and contraction with the modulation of the scaling factor, which changes the affiliation function without adding fuzzy rules. The fuzzy rule tables for the basis fuzzy controller and the scaling factor fuzzy controller are shown in Tables 3 and 4, respectively.

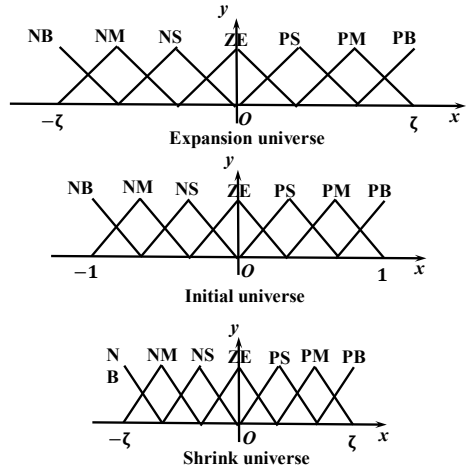

Fig. 8. Variable universe fuzzy control.

**Table 3** Fuzzy rules for the basis fuzzy controller

|       | $y_d$ | $ZMP_{ec}$ | | | | | | |
|-------|-------|----|----|----|----|----|----|----|
|       |       | NB | NM | NS | ZO | PS | PM | PB |
|       | NB    | PB | PB | PB | PM | PM | PS | PS |
|       | NM    | PB | PB | PM | PM | PS | PS | PS |
|       | NS    | PM | PM | PS | PS | PS | PS | PS |
| $ZMP_e$ | ZO  | PM | PS | PS | ZO | NS | NS | NM |
|       | PS    | NS | NS | NM | NM | NM | NM | NB |
|       | PM    | NS | NM | NM | NB | NB | NB | NB |
|       | PB    | NS | NM | NB | NB | NB | NB | NB |

**Table 4** Fuzzy rule for scaling factor fuzzy controller

| scaling factor | Stability Level | | | | |
|----------------|-----|-----|-----|-----|-----|
|                | $L_1$ | $L_2$ | $L_3$ | $L_4$ | $L_5$ |
| $\zeta(ZMP_e)$ | VS | S | M | B | VB |
| $\xi(ZMP_{ec})$ | VS | S | M | B | VB |
| $\gamma(y_d)$ | VS | S | M | B | VB |

3) Design of active disturbance rejection control

The stability of the unmanned deformable vehicle in the humanoid walking process is controlled by the X- and Y-direction adjusting COM sliders in the COM adjustment mechanism. In the humanoid walking motion, the support domain is small and the position of the COM of the system changes significantly. At this time, the unmanned deformable vehicle is prone to tipping, and the walking stability is very sensitive to external interference. Therefore, the ADRC with a strong anti-interference is used to control the walking stability. Moreover, the electromechanical model of the control object COM adjustment mechanism is a simpler linear second-order system, which makes it easier to establish the expanded state observer of the ADRC.

*a) Linear Expanded State Observer:* From Eq. (4), the electromechanical model of both the X- and Y-direction COM adjustment mechanisms is a linear second-order system. $f$ is defined as the total system disturbance, which is differentiable, $f = -B_z \dot{y}/J$. $b_0$ is defined as the input coefficient of the control system, $b_0 = K_a K_t r_g/J$. Thus, Eq. (4) can be rewritten as

$$\ddot{y} = f + b_0 u, \tag{19}$$

where $y$ and $u$ are the output and input quantities of the controlled object, respectively.

The state variables are defined as $x_1 = y$, $x_2 = \dot{y}$, $x_3 = f$, and $h = \dot{f}$. The expanded state expression for the electromechanical model is

$$\begin{cases} \dot{x} = Ax + Bu + Eh \\ y = Cx \end{cases}, \tag{20}$$

where $x = \begin{bmatrix} x_1 \\ x_2 \\ x_3 \end{bmatrix}$, $A = \begin{bmatrix} 0 & 1 & 0 \\ 0 & 0 & 1 \\ 0 & 0 & 0 \end{bmatrix}$, $B = \begin{bmatrix} 0 \\ b_0 \\ 0 \end{bmatrix}$, $E = \begin{bmatrix} 0 \\ 0 \\ 1 \end{bmatrix}$, and $C = [1 \ 0 \ 0]$.

A Luenberger observer is established as follows:

$$\begin{cases} \dot{z} = Az + Bu + L(y - \hat{y}) \\ \hat{y} = Cz \end{cases}, \tag{21}$$

where $z = [z_1 \ z_2 \ z_3]^T$ is the observation state vector, $z_i \to x_i, (i = 1,2,3)$, and $L = [\varphi_1 \ \varphi_2 \ \varphi_3]^T$ is the observation error feedback gain matrix.

Defining the observation error $e_i = x_i - z_i$ and subtracting Eq. (20) from Eq. (21) yields

$$\dot{e} = \dot{x} - \dot{z} = Ax + Bu - Az - Bu - LC(x - z) + Eh$$
$$= (A - LC)e + Eh. \tag{22}$$

From modern control theory [25], it is known that the observation error $e \to 0$ is conditional on the fact that the eigenvalues of the matrix $A - LC$ have a negative real part. The expression of the matrix $A - LC$ is given by

$$A - LC = \begin{bmatrix} -\varphi_1 & 1 & 0 \\ -\varphi_2 & 0 & 1 \\ -\varphi_3 & 0 & 0 \end{bmatrix}. \tag{23}$$

The characteristic polynomial of the matrix $A - LC$ is

$$|\lambda I - (A - LC)| = (\lambda + \varphi_1)\lambda^2 + \varphi_2 \lambda + \varphi_3$$
$$= \lambda^3 + \varphi_1 \lambda^2 + \varphi_2 \lambda + \varphi_3. \tag{24}$$

According to the bandwidth method, the observer bandwidth is defined as $\omega_o$, and all three poles of the observer are configured at $-\omega_o$ to obtain:

$$|\lambda I - (A - LC)| = (\lambda + \omega_0)^3 = \lambda^3 + \varphi_1 \lambda^2 + \varphi_2 \lambda + \varphi_3. \tag{25}$$

From Eq. (25), the observer error feedback gain is given by

$$\begin{cases} \varphi_1 = 3\omega_o \\ \varphi_2 = 3\omega_o^2 \\ \varphi_3 = \omega_o^3 \end{cases}. \tag{26}$$

From the above analysis, the matrix $A - LC$ is stable and $z \to x$ as long as $\omega_o > 0$.

*b) Fastest Tracking Differentiator:* In the process of stability adjustment, the desired displacement $y_d$ of the adjusting COM slider output by the fuzzy controller is a very complex nonlinear curve, and a large overshoot occurs in the process of fast tracking it. In order to suppress the overshoot of the control, the fastest tracking differentiator is used to smooth the desired displacement, which can well balance the contradiction between overshoot and rapidity, as well as enhance the robustness of the system and the noise suppression ability [26]. The discrete form of the fastest tracking differentiator is

$$\begin{cases} r_{d1}(k+1) = r_{d1}(k) + \Delta t \cdot r_{d2}(k) \\ r_{d2}(k+1) = r_{d2}(k) + \Delta t \cdot fst(r_{d1} - y_d, r_{d2}, r, h_0) \end{cases}, \tag{27}$$

$$\begin{cases} p = rh_0 \\ p_0 = h_0 p \\ v = r_{d1} + h_0 r_{d2} \\ w_0 = \sqrt{p^2 + 8r|v|} \\ w = \begin{cases} r_{d2} + \dfrac{w_0 - p}{2}, |v| > p_0 \\ r_{d2} + \dfrac{v}{h_0}, |v| \le p_0 \end{cases} \\ fst = -\begin{cases} rsgn(w), |w| > p \\ r\dfrac{w}{p}, |w| \le p \end{cases} \end{cases}, \tag{28}$$

where $r_{d1}$ is the tracking signal of the input $y_d$, $r_{d2}$ is the approximate differential signal of $y_d$, $\Delta t$ is the discrete sampling period, $r$ is the speed factor, which determines the rate of convergence of the fastest tracking differentiator, and $h_0$ is the filtering factor, which acts as a filter on the noise, and $sgn(x)$ is the sign function.

*c) State Error Feedback:* ADRC uses error feedback control to actively compensate for the total disturbance estimated by the state observer, which not only achieves the effect of anti-disturbance but also transforms the controlled object into an integral series-type structure. The input control $u_o$ of the electromechanical model can be designed as

$$\begin{cases} u = k_p(r_{d1} - z_1) + k_d(r_{d2} - z_2) \\ u_o = \dfrac{u - z_3}{b_0} \end{cases}, \tag{29}$$

where $k_p$ and $k_d$ are proportional and differential control parameters, respectively, and $u$ is the output control physical quantity of the state error feedback controller.

With a reasonable observer design, the following conclusions can be drawn:

$$z_1 \to y, z_2 \to \dot{y}, z_3 \to f. \tag{30}$$

By combining Eq. (29) and Eq. (30), the output $u$ of the state error feedback control law can be represented as

$$\begin{aligned} u &= k_p(r_{d1} - z_1) + k_d(r_{d2} - z_2) \\ &= k_p(r_{d1} - y) + k_d(r_{d2} - \dot{y}). \end{aligned} \tag{31}$$

Applying the Laplace transform to Eq. (31) yields

$$U(s) = k_p\big(R(s) - Y(s)\big) + k_d\big(sR(s) - sY(s)\big). \tag{32}$$

By combining Eq. (19) and Eq. (29), it can be determined that:

$$\ddot{y} = u. \tag{33}$$

Applying the Laplace transform to Eq. (33) yields

$$s^2 Y(s) = U(s). \tag{34}$$

Combining Eq. (32) and Eq. (34) yields

$$\frac{Y(s)}{R(s)} = \frac{k_d s + k_p}{s^2 + k_d s + k_p}. \tag{35}$$

By placing the poles of Eq. (35) at the frequency $\omega_c$, the following expression is obtained:

$$s^2 + k_d s + k_p = (s + \omega_c)^2. \tag{36}$$

The final control parameters are obtained as

$$k_p = \omega_c^2, k_d = 2\omega_c. \tag{37}$$

## 5. Simulation of multi-configuration motion stability control based on COM adjusting mechanism

*5.1. Stability control for steady-state steering*

When a vehicle is driving with steady-state steering, in order to ensure a good lateral stability, the vehicle needs to have a moderate understeering capability (stability factor K slightly greater than zero). Under the condition that the vehicle structure and parameters are determined, its stability factor cannot be changed actively. The unmanned deformable vehicle can control the displacement of the X-direction adjusting COM slider, thus changing the front and rear wheelbases of the vehicle and the stability factor K, so that the vehicle can have better lateral stability under different steering conditions. Under the condition of a lateral acceleration of 0.3g, the control objective was the desired stability factor $K_d = 0.0024$ [27], and the steady-state steering stability control of the vehicular state was simulated and verified according to the control strategy designed above.

**Table 5** Performance indicators of the control system

| control methods | steady-state error /(%) | overshoot /(%) | rising time /(s) | adjusting time /(s) |
|---|---|---|---|---|
| PID | 0.11 | 14.99 | 1.05 | 2.37 |
| Fuzzy-PID | 0.09 | 9.19 | 0.35 | 1.52 |

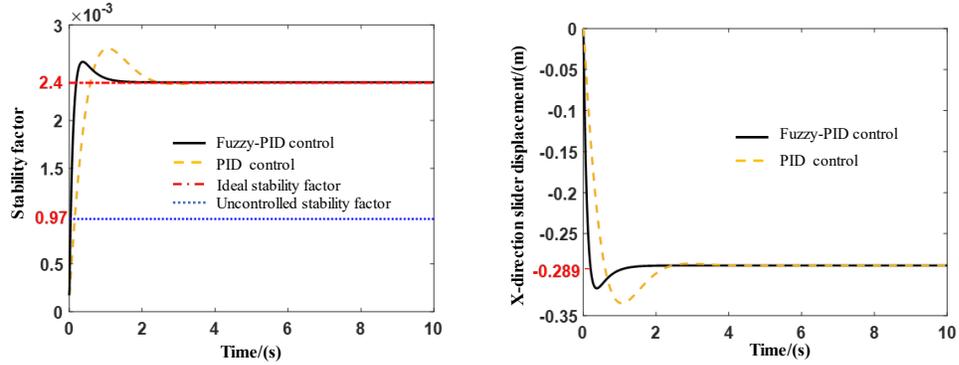

(a) changes in stability factor          (b) X-axis adjustment of COM slider displacement
**Fig. 9.** Simulation results of steady-state steering driving stability control.

As shown in Fig. 9(a), the slider was located at the center of the slide rail and remained motionless before stability control was performed. Since the position of the COM of the vehicle did not change, the stability factor K was always 0.00097. When stability control was performed, both methods, PID control and Fuzzy-PID control, were able to ensure that the stability factor curve converged to the desired value of 0.0024. However, the three control coefficients of the Fuzzy-PID control were variable, so the Fuzzy-PID had a better control effect than the PID control. A comparison of the control performance indices of the PID and Fuzzy-PID is shown in Table 5. As shown in Fig. 9(b), by controlling the COM adjustment mechanism, the slider moved 0.289 m in the negative direction of the X-axis, which in turn led to a change of the COM position of the vehicle. Based on the steering dynamics equation, it was calculated that the stability factor K was controlled to 0.0024.

*5.2. Walking stability control*

The parameters of the X-direction ADRC were $\omega_0 = 1000$, $\omega_c = 200$, and $b_0 = 0.08$. The parameters of the Y-direction ADRC were $\omega_0 = 1200$, $\omega_c = 250$, and $b_0 = 0.08$. The VUFC-ADRC stability control simulation was performed and analyzed for the first 6 s of the walking process.

As shown in Fig. 10(a), the PID control algorithm not only underwent a greater control deviation when tracking the desired trajectory of the X-direction ZMP but also had a poor tracking effect, and the maximum tracking error reached 0.037 m. Therefore, the walking stability margin of the unmanned deformable vehicle was insufficient under the PID stability control algorithm, and it was prone to tipping and destabilizing under the external disturbances. In contrast, both VUFC-ADRC control algorithms (with and without the K-means clustering algorithm) could track the desired ZMP more quickly and smoothly and show a better control effect. However, during the transition between single-legged and double-legged support periods, after quantitative preprocessing of the stability level by the K-means clustering algorithm, the overshoot of the VUFC-ADRC control algorithm was reduced by 14.86%, and the steady-state error was also reduced by 3.6%. Therefore, the K-means clustering algorithm could not only more accurately quantify the stability changes during the walking process but could also further improve the control effect of the VUFC-ADRC. Fig. 10(b) also exhibited similar control results, where the PID had the worst control effect with large control deviations when tracking the desired trajectory of the Y-direction ZMP, while the VUFC-ADRC with the K-means clustering algorithm exhibited the least overshoot and the lowest steady-state error. Therefore, all three control strategies could ensure the stability of the unmanned deformable vehicle during the walking process, but the VUFC-ADRC optimized with K-means clustering performed the best in controlling the ZMP, which ensured the optimal walking stability.

As shown in Figs. 10(c) and 10(d), the walking stability levels of the unmanned deformable vehicle under the control of the two VUFC-ADRC algorithms were generally low, basically remaining at levels of 1 and 2. The VUFC-ADRC control without K-means clustering optimization showed more significant deviation during the switching between the single-legged and double-legged support periods, and it even reached rank 3 at some moments. This was due to the faster change of the desired ZMP during the switching period, which led to a higher requirement of the system to regulate the slider in real time, thus deteriorating the stability level.

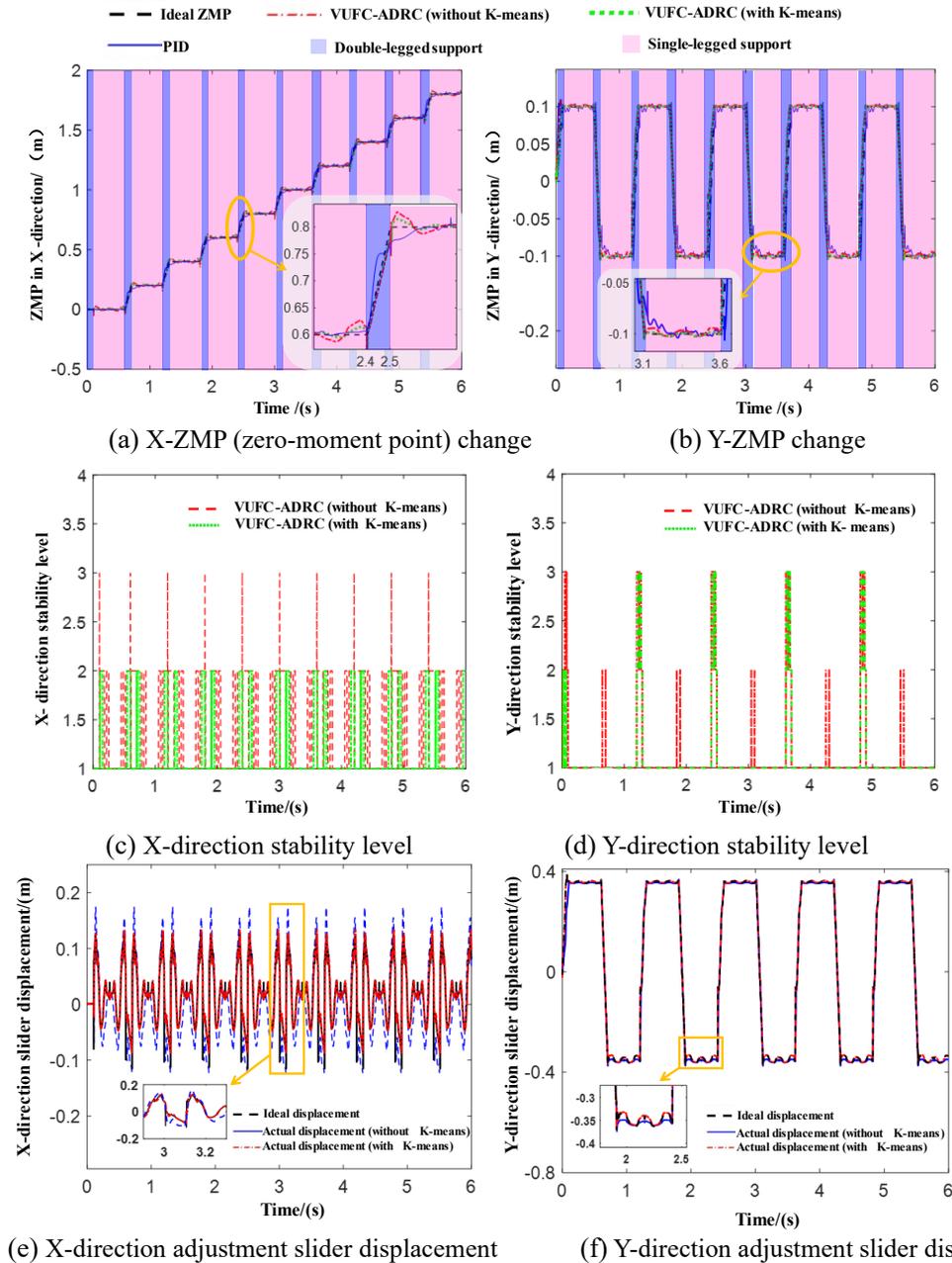

(a) X-ZMP (zero-moment point) change  (b) Y-ZMP change
(c) X-direction stability level  (d) Y-direction stability level
(e) X-direction adjustment slider displacement  (f) Y-direction adjustment slider displacement
**Fig. 10.** Simulation results of walking stability control.

As shown in Figs. 10(e) and 10(f), there was a deviation and disturbance in the expected displacement of the slider during the alternation of the single-legged and double-legged support periods, and both VUFC-ADRC algorithms were able to overcome this deviation and disturbance effectively, which proved that the designed VUFC-ADRC exhibited a good anti-disturbance ability in the walking stability control algorithm. The VUFC-ADRC optimized by K-means clustering even reduced the maximum overshoot by 23% compared to the VUFC-ADRC without K-means clustering optimization when tracking the desired displacement of the slider. Thus, the quantification of the stability level using the K-means clustering algorithm could optimize the tracking effect of the VUFC-ADRC on the desired displacement of the slider.

## 6. Test of multi-configuration motion stability control based on COM adjusting mechanism

### 6.1. Construction of test platform

In order to verify the effectiveness of the stability control strategy based on the COM adjustment mechanism, the COM adjustment mechanism was arranged on the unmanned deformable vehicle, and the multi-configuration motion stability control test platform was constructed, as shown in Fig. 11.

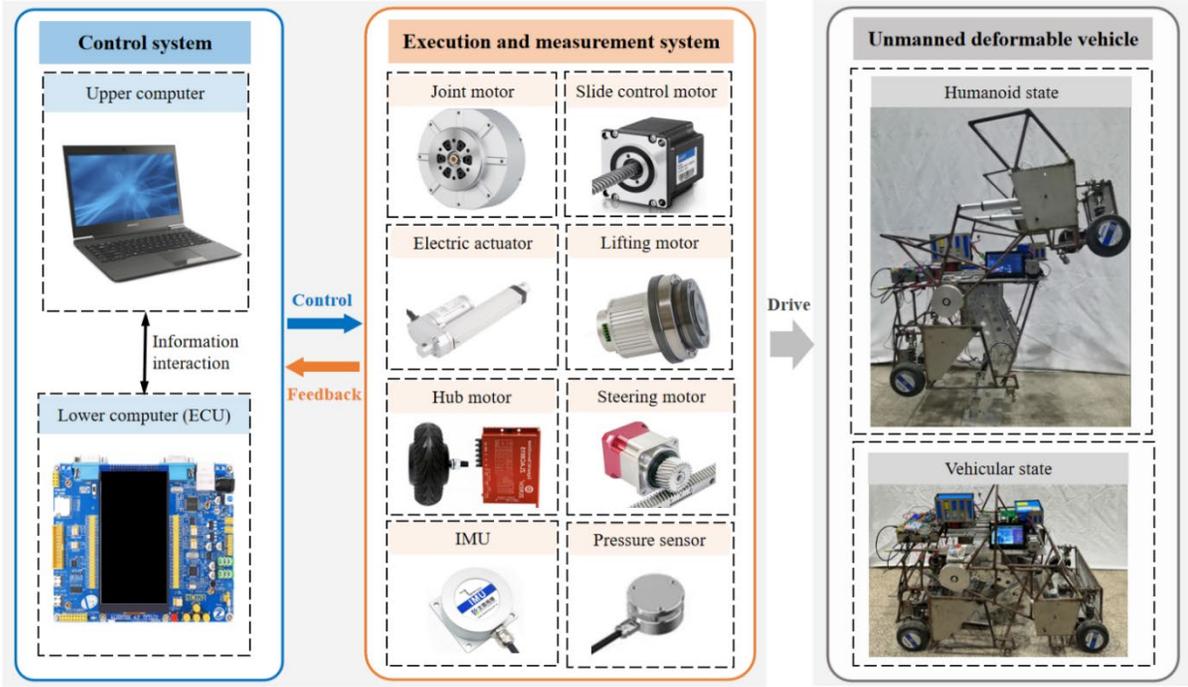

**Fig. 11.** Unmanned deformable vehicle test platform.

The test platform consisted of an unmanned deformable vehicle prototype, an actuation and measurement system, and a control system. In the actuation system, joint motors, wheel hub motors, and steering motors were used to drive the unmanned deformable vehicle to walk in the humanoid state and to steer in the vehicular state, respectively. The lifting motor and electric actuator were responsible for lifting the front body to assist in accomplishing the configuration transition between the vehicular state and the humanoid state. The slide control motor in the COM adjustment mechanism drove the movement of the slider, which in turn controlled the stability of the multi-configuration motion of the unmanned deformable vehicle. The measurement system included encoders, an inertial measurement unit, and pressure sensors. The encoder in the hub motor was used to measure the driving speed of the vehicular state. The inertial measurement unit (IMU, BW-IMU200) was placed on the bottom plate of the rear body and was used to collect the yaw rate signals of the vehicular state during steady-state steering. A pressure sensor (HZC-30B) was installed at each of the four corner positions of each plantar foot to measure the change of the center-of-pressure (COP) during the walking process, and the ZMP was approximated by the COP [28]. The lower computer in the control system judged the stability state of the unmanned deformable vehicle based on the signals collected by the sensors and then selected the corresponding control algorithm to output the control voltage to the slider control motor. The multi-configuration motion stability control was realized by changing the displacements of the X- and Y-direction sliders. The upper computer in the control system interacted with the lower computer through a communication interface to monitor the motion state of the unmanned deformable vehicle in real time.

*6.2. Stability control test for steady-state steering*

Based on the fixed front wheel steering angle method [29], stability control tests were conducted for steady-state steering of the vehicular state. The steering angle of the front wheels was changed by a steering motor. In the test, the unmanned deformable vehicle was driven around a circle with a radius $R_0$ of 15 m at the lowest stable speed. When the circumferential steering reached stability, the steering angle $\delta$, the vehicle's speed $u_x$, and the yaw rate $\omega_r$ were measured through sensors. While keeping the front wheel angle unchanged, the driving speed of circular steering was increased, and when the circular driving condition reached stability, the vehicle speed $u_x$ and the yaw rate $\omega_r$ were measured again. The lateral acceleration $a_y$ was increased by no more than 0.5 m/s² each time, and the above test was repeated until $a_y$ reached 5 m/s².

Three tests were performed for both left and right steering, and the formula for calculating the steering radius $R$ at each point is as follows:

$$R = \frac{u_x}{\omega_r}. \tag{38}$$

The lateral acceleration $a_y$ is given by

$$a_y = u_x \omega_r. \tag{39}$$

The deviation between the front wheel slip angle and the rear wheel slip angle in the steady-state steering test is

$$\alpha_1 - \alpha_2 = 57.3 L \left( \frac{1}{R_0} - \frac{1}{R} \right). \tag{40}$$

The relationship between the stability factor and the slip angle of the front and rear wheels at small lateral accelerations is as follows [30]:

$$K = \frac{\alpha_1 - \alpha_2}{a_y L}. \tag{41}$$

The deviation between the front wheel slip angle and the rear wheel slip angle is also an important index to measure the steady-state steering performance of the vehicle. In the process of vehicle steering, increasing the front wheel slip angle can weaken the steering tendency, while increasing the rear wheel slip angle can enhance the steering tendency, so the vehicle has an understeering tendency when the deviation between the front wheel and the rear wheel slip angle is greater than zero. In the test process, the trend of the deviation between the front and rear wheel slip angles with the lateral acceleration is shown in Fig. 12(a). Before and after the steady-state steering control, the deviation between the front and rear wheel slip angle was greater than 0, the vehicular state had an understeering tendency. When the lateral acceleration was below 0.3g, $\alpha_1 - \alpha_2$ and $a_y$ were basically linear. However, after the lateral acceleration exceeded 0.3g, the relationship between $\alpha_1 - \alpha_2$ and $a_y$ became nonlinear. This was because the tire exhibited nonlinear dynamics after the lateral acceleration was greater than 0.3g [31]. Before the lateral acceleration of 0.3g, calculations with (41) showed that the stability factor K after the steady-state steering control basically stayed near the target value of 0.0024, while the stability factor K before control was 0.00097. Fig. 12(b) shows the graph of the yaw rate gain with the longitudinal vehicle speed. Before and after the steady-state steering stability control, the yaw rate gain curves were all located under the neutral steering straight line, i.e., they all had understeering characteristics. However, the characteristic vehicle speed $u_{ch}$ without steady-state steering stability control exceeded 105 km/h, the yaw rate gain continued to increase with the increase in the vehicle speed, and the steady-state steering stability became worse and worse with the increase in the vehicle speed. After the steady-state steering stability control, the characteristic vehicle speed $u_{ch}$ of the vehicular state was about 75 km/h, which meant that the yaw rate gain curve tended to stabilize when the vehicle speed exceeded 75 km/h. Therefore, the steady-state steering stability control could further enhance the understeering characteristics of the vehicle and improve the ability to resist lateral instability during high-speed steering.

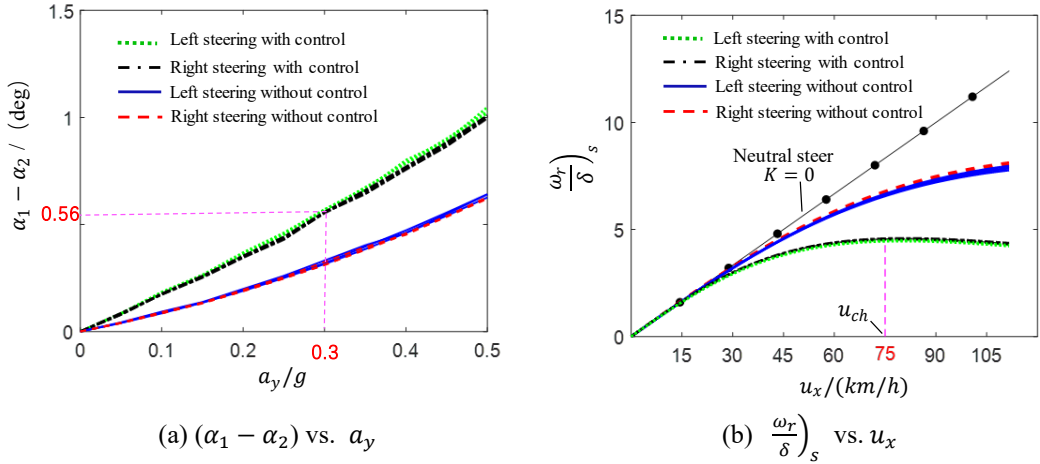

(a) $(\alpha_1 - \alpha_2)$ vs. $a_y$

(b) $\left.\frac{\omega_r}{\delta}\right)_s$ vs. $u_x$

Fig. 12. Steady-state response curves of the vehicle.

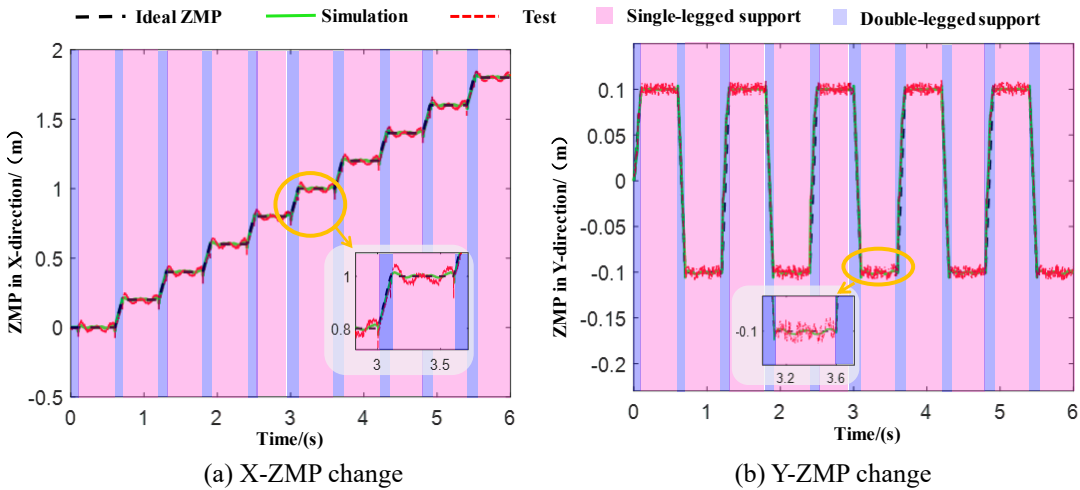

(a) X-ZMP change

(b) Y-ZMP change

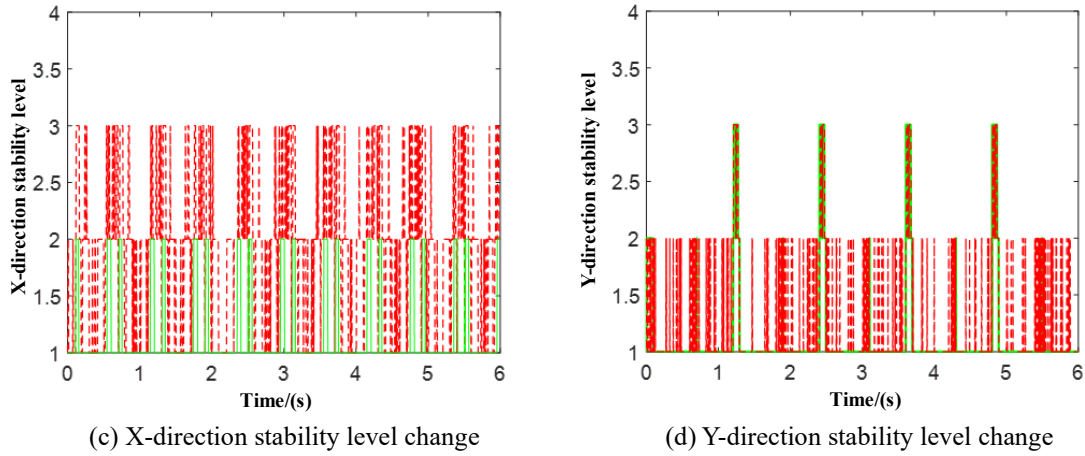

(c) X-direction stability level change      (d) Y-direction stability level change

**Fig. 13.** Test results of walking stability control.

*6.3. Controlled test of walking stability*

According to the robot walking gait planning, the walking test of the unmanned deformable vehicle humanoid state was conducted on a flat road with a gait period of 1.2 s (two single-legged support periods of 0.5 s × 2 and two double-legged support periods of 0.1 s × 2). From Figs. 13(a) and 13(b), it can be seen that there was basically no significant deviation between the X- and Y-direction ZMP curves in the walking test and the corresponding simulated ZMP curves. They were all well controlled in the vicinity of the ideal ZMP curves, which indicated that the designed VUFC-ADRC stability control algorithm was able to effectively realize the expected stability control effect during actual walking. In addition, Figs. 13(c) and 13(d) show the X- and Y-direction stability levels during walking, respectively. The X-direction stability level mainly fluctuated up and down in the 2nd level, indicating that the X-direction walking stability was better. However, in the actual walking scene, the control system needed to deal with more complex external disturbances, so the X-direction stability level in the test was overall one level higher than that in the simulation results. The walking stability level in the Y-direction was significantly better than that in the X-direction, which was basically below level 2. Under the influence of complex interference in the test, the stability level of the Y-direction in the test was also one level higher than the simulation results. In summary, through the designed walking stability control algorithm, under the action of the COM adjustment mechanism, the unmanned deformable vehicle humanoid state showed good stability in the walking process. Although the test stability level will be higher than the simulation result due to the influence of external interference in the actual test, it can still meet the expected walking stability requirements.

## 7. Conclusion

To address the motion stability challenges of an unmanned deformable vehicle with multiple reconfigurations (vehicular state and humanoid state), a novel COM adjustment mechanism was designed. Using hybrid automata modeling, a hierarchical stability control system was implemented to manage the vehicle's motion across all configurations with the COM adjusting mechanism. The upper-level decision controller assessed the vehicle's configuration and motion based on real-time system parameter, selecting the appropriate control strategy. The lower-level controller utilized Fuzzy-PID and VUFC-ADRC algorithms for targeted stability control: steady-state steering in vehicular mode and walking stability in humanoid mode.

The longitudinal positioning of a vehicle's COM considerably impacted steady-state steering stability. The longitudinal COM position of the vehicle was not able to be controlled by conventional vehicles, resulting in a possible failure to satisfy steady-state steering stability in both unloaded and fully loaded conditions. Through the Fuzzy-PID control of the slider displacement in the COM adjusting mechanism, the stability factor was maintained near the optimal value of 0.0024 under various working conditions. The unmanned deformable vehicle had been able to maintain better understeer characteristics, and its steady-state steering stability was significantly better than that of the conventional vehicle.

To address the walking stability control in humanoid state, the VUFC-ADRC control algorithm was designed based on the COM adjustment mechanism, and quantitatively rated the stability level by K-means clustering algorithm. The control results showed that after the K-means clustering algorithm quantified the stability level, the algorithm reduced the overshooting amount by 14.86%, the steady-state error by 3.6%, and effectively eliminated the internal disturbances and suppressed mutations. Under the control of the COM adjustment mechanism, the walking stability level of the unmanned deformable vehicle in the humanoid state was basically maintained below the 3rd level, thus significantly improving its walking stability and anti-disturbance ability.

**Declaration of Conflicting Interests**

**Funding**

This work was supported by the National Natural Science Foundation of China (Grant No. 51875148); and the Key Technologies Research and Development Program of Anhui Province (Grant No. 202104a05020040).

**Acknowledgments**

We thank LetPub (www.letpub.com) for its linguistic assistance during the preparation of this manuscript.

**References**

[1] J. Liu, D. Zhao, C. Liu, et al., "Reconfiguration motion analysis and motion quality control of an unmanned metamorphic vehicle," Control. Eng. Pract., vol.142, pp.105776, 2024, doi: https://doi.org/10.1016/j.conengprac.2023.105776.
[2] J. Liu, J. Zhu, D. Zhao, et al., "Integrated optimization design and motion control of multi-configuration unmanned metamorphic vehicle," Adv. Eng. Inf., vol. 59, pp.102325, 2024, doi: https://doi.org/10.1016/j.aei.2023.102325.
[3] J. Liu, J. Xing, C. Liu, et al., "Reconfiguration of stability and control strategy for some unmanned deformable vehicles," Appl. Math. Model., 2024, vol. 127, pp. 784-802, doi: https://doi.org/10.1016/j.apm.2024.01.005.
[4] J. Liu, J. Song, H. Li, et al., "Direct yaw-moment control of vehicles based on phase plane analysis," Proc. Inst. Mech. Eng. Part D: J. Automob. Eng., vol. 236, pp. 2459-2474, 2022, doi: https://doi.org/10.1177/09544070211052375.
[5] J. Liu, Z. Yan, M. Lu, et al., "Stability analysis and design of an unmanned deformable vehicle during coupled reconfiguration motion," Proc. Inst. Mech. Eng. Part K: J. Multi-body Dyn., vol. 236, pp. 639-659, 2022, doi: https://doi.org/10.1177/1464419322112.
[6] J. Liu, X. Ruan, M. Lu, et al., "Motion analysis and stability optimization for metamorphic robot reconfiguration," Robot., vol. 41, no. 4, pp. 1179-1202, 2023, doi: https://doi.org/10.1017/S0263574722001618.
[7] K. Hu, C. Ott, D. Lee, "Learning and Generalization of Compensative Zero-Moment Point Trajectory for Biped Walking," IEEE Trans. Robot., vol. 32, no. 3, pp. 717-725, 2016, doi: https://doi.org/10.1109/TRO.2016.2553677.
[8] S. Caron, Q. C. Pham, Y. Nakamura, "ZMP Support Areas for Multicontact Mobility Under Frictional Constraints," IEEE Trans. Robot., vol. 33, no. 1, pp. 67-80, 2017, doi: https://doi.org/10.1109/TRO.2016.2623338.
[9] N. Scianca, F.M. Smaldone, L. Lanari, et al., "A feasibility-driven MPC scheme for robust gait generation in humanoids," Robot. Auton. Syst., vol. 189, p. 104957, 2025, https://doi.org/10.1016/j.robot.2025.104957.
[10] K. Zhou, Ch. Li, C. Li, "A Motion Planning Method for Quadrupedal Robots for Unknown Complex Terrain," J. Mech. Eng., vol. 56, no.2, pp. 210-219, 2020.
[11] J. Yi, Q. Zhu, J, Wu. "Optimal control-based vibration suppression for walking humanoid robots," Robot, vol. 40, no.2, pp. 129-135, 2018, doi: https://doi.org/10.13973/j.cnki.robot.170308.
[12] S. Kim, Y. J. Han, Y. D. Hong, "Stability control for dynamic walking of bipedal robot with real-time capture point trajectory optimization," J. Intell. Robot. Syst., 2019, vol. 96, no. 3, pp. 345-361, 2019, doi: https://doi.org/10.1007/s10846-018-0965-7.
[13] H. K. Shin, B. K. Kim, "Energy-Efficient Gait Planning and Control for Biped Robots Utilizing the Allowable ZMP Region," IEEE Trans. Robot., vol. 30, no. 4, pp. 986-993, 2014, doi: https://doi.org/10.1109/TRO.2014.2305792.
[14] N. Scianca, D. De Simone, L. Lanari, et al., "MPC for Humanoid Gait Generation: Stability and Feasibility," IEEE Trans. Robot., vol. 36, no. 4, pp. 1171-1188, 2020, doi: https://doi.org/10.1109/TRO.2019.2958483.
[15] M. Iida, H. Nakashima, H. Tomiyama, et al, "Small-radius turning performance of an articulated vehicle by direct yaw moment control," Comp. Electron. Agric., vol. 76, no. 2, pp. 277-283, 2011, doi: https://doi.org/10.1016/j.compag.2011.02.006.
[16] M. Gözü, B. Ozkan, M. T. Emirler, "Disturbance observer based active independent front steering control for improving vehicle yaw stability and tire utilization," Int. J. Autom. Technol., vol. 23, np. 3, pp. 841-854, 2022, doi: https://doi.org/10.1007/s12239-022-0075-1.
[17] M. Jalali, S. Khosravani, A. Khajepour, et al., "Model predictive control of vehicle stability using coordinated active steering and differential brakes," Mechatron., vol. 48, pp. 30-41, 2017, doi: https://doi.org/10.1016/j.mechatronics.2017.10.003.
[18] Z. Yu, "Automobile Theory," Bei Jing, China, China Machine Press, 2007.
[19] K. Wang, Z. J. Hu, P. Tisnikar, et al., "When and where to step: Terrain-aware real-time footstep location and timing optimization for bipedal robots," Robot. Auton. Syst., vol. 179, p. 104742, 2024, doi: https://doi.org/10.1016/j.robot.2024.104742.
[20] E. Pecker-Marcosig, S. Zudaire, R. Castro, et al., "Correct and efficient UAV missions based on temporal planning and in-flight hybrid simulations," Robot. Auton. Syst., vol. 164, p. 104404, 2023, doi: https://doi.org/10.1016/j.robot.2023.104404.
[21] Z. Chen, S. Xun, "Hybrid automaton-based disturbance-aware predictive control with receding horizon optimization for three-phase full-bridge inverters," Control. Eng. Pract., vol. 121, p. 104984, 2022, doi: https://doi.org/ 10.1016/j.conengprac.2021.104984.
[22] X. Gu, K. Meng, X. Yao, "Combinatorial clustering based lateral stability discrimination method for intelligent vehicles," Autom. Eng., vol. 42, no. 11, pp. 1497-1505, 2020, doi: https://doi.org/10.19562/j.chinasae.qcgc.2020.11.007.
[23] X. Wang, Z. Sun, D. He, et al., "Incremental fast relevance vector regression model based multi-pollutant emission prediction of biomass cogeneration systems," Control. Eng. Pract., vol. 149, p. 105986, 2024, doi: https://doi.org/10.1016/j.conengprac.2024.105986.
[24] R. H. Du, H. F. Hu, K. Gao, "Study on trajectory tracking control of self-driving car based on variable predictive time domain MPC," J. Mech. Eng., vol. 58, no. 24, pp. 275-288, 2022, doi: https://doi.org/10.3901/JME.2022.24.275.
[25] G. Zhao, "Modern Control Theory," Bei Jing, China, Machinery Industry Press, 2018.
[26] Q. Han, "From PID technology to "self-immobilizing control" technology," Control Eng., vol. 03, pp. 13-18, 2002, doi: https://doi.org/10.3969/j.issn.1671-7848.2002.03.003.
[27] D. Karnopp, "Vehicle stability," New York, USA, Marcel Dekker, 2004.
[28] GB/T6323.6-1994, "Test Method for Automobile Handling Stability: Steady State Slewing Test," Bei Jing, China, China Standard Press, 1994.
[29] Y. de Viragh, M. Bjelonic, C. D. Bellicoso, et al., "Trajectory Optimization for Wheeled-Legged Quadrupedal Robots Using Linearized ZMP Constraints," IEEE Robot. Autom. Lett., vol.4, no. 2, pp. 1633-1640, 2019, doi: https://doi.org/10.1109/LRA.2019.2896721.
[30] Yang, D. Dong, C. Ma, et al., "Stability control for electric vehicles with four in-wheel-motors based on sideslip angle," World Electr. Veh. J., vol.12, no.1, pp. 42-57, 2021, doi: https://doi.org/10.3390/wevj12010042.
[31] X. Sun, H. Zhang, Y. Cai, et al., "Hybrid modeling and predictive control of intelligent vehicle longitudinal velocity considering nonlinear tire dynamics," Nonlinear Dyn., vol.97, pp. 1051-1066, 2019, doi: https://doi.org/10.1007/s11071-019-05030-5.